\documentclass[letterpaper, 10 pt, conference]{ieeeconf}
\IEEEoverridecommandlockouts
\usepackage{balance}
\usepackage{amsmath} 
\usepackage{amssymb}  

\usepackage{amsthm}
\usepackage{xspace}
\usepackage[linesnumbered,ruled,noend]{algorithm2e}

\usepackage{xcolor}
\definecolor{gk}{RGB}{120, 120, 120}
\definecolor{gg}{HTML}{5f9411}
\definecolor{gb}{HTML}{417598}
\definecolor{gr}{HTML}{d15120}
\definecolor{gy}{HTML}{d2ad00}

\usepackage{graphicx}
\usepackage{svg}
\let\labelindent\relax
\usepackage[inline]{enumitem}

\usepackage[
activate   = {true},
protrusion = false,
expansion  = true,
kerning    = true,
spacing    = true,
tracking   = false,
auto       = true,
selected   = true,
factor     = 1000,
stretch    = 15,
shrink     = 15,
]{microtype}

\usepackage{csquotes}
\usepackage[
maxbibnames=99,
maxcitenames=2,
natbib=true,
style=numeric-comp,
backend=biber,
sorting=none,
giveninits=true,
url=false,
doi=false,
isbn=false,
]{biblatex}

\makeatletter
\let\NAT@parse\undefined
\makeatother
\usepackage[pdfa,colorlinks,bookmarksopen,bookmarksnumbered,allcolors=gg]{hyperref}
\usepackage[english]{babel}

\usepackage{xurl}
\usepackage[nameinlink,capitalise]{cleveref}
\usepackage{tabularx}
\usepackage{adjustbox}
\usepackage{array, booktabs, makecell}

\usepackage{draftwatermark}

\SetWatermarkText{Appears in IEEE International Conference on Robotics and Automation (ICRA) 2023}
\SetWatermarkColor[gray]{0.5}
\SetWatermarkFontSize{0.3cm}
\SetWatermarkAngle{0}
\SetWatermarkHorCenter{2.79in}
\SetWatermarkVerCenter{0.4in}

\newcommand{\transit}[1]{\texttt{TRANSIT(}{#1}\texttt{)}}
\newcommand{\transito}{\texttt{TRANSIT}\xspace}
\newcommand{\transfer}[1]{\texttt{TRANSFER(}{#1}\texttt{)}}
\newcommand{\transfero}{\texttt{TRANSFER}\xspace}
\newcommand{\rrtconnect}{\texttt{RRTConnect}\xspace}
\newcommand{\SE}[1]{\ensuremath{\operatorname{SE}(#1)}}
\newcommand{\arrangement}{\ensuremath{\alpha}}
\newcommand{\arrangementop}[1]{\ensuremath{\operatorname{\alpha}\lbrack#1\rbrack}}
\newcommand{\Arrangements}{\ensuremath{\mathcal{A}}}
\newcommand{\object}{\ensuremath{o}}
\newcommand{\Objects}{\ensuremath{\mathcal{O}}}
\newcommand{\sethree}{\SE{3}}

\newcommand{\True}{\texttt{True}}
\newcommand{\False}{\texttt{False}}
\newcommand{\collision}[1]{\ensuremath{\operatorname{\texttt{CollisionFree}}(#1)}}
\newcommand{\stable}[1]{\ensuremath{\operatorname{\texttt{Stable}}(#1)}}
\newcommand{\config}{\ensuremath{q}}
\newcommand{\CSpace}{\ensuremath{\mathcal{Q}}}

\newcommand{\robot}{\ensuremath{r}}

\let\oldnl\nl
\newcommand{\nonl}{\renewcommand{\nl}{\let\nl\oldnl}}
\newcommand{\myTitle}{%
Object Reconfiguration with Simulation-Derived Feasible Actions%
}
\hypersetup{%
    pdftitle={\myTitle{}},
    pdfauthor={Yiyuan Lee, Wil Thomason, Zachary Kingston, Lydia E. Kavraki},
    pdfkeywords={TODO},
    pdfsubject={TODO},
    colorlinks=true,allcolors=gr,
}%
\newcolumntype{Y}{>{\centering\arraybackslash}X}

\crefname{line}{line}{lines}
\crefname{figure}{Fig.}{Figs.}
\Crefname{figure}{Fig.}{Figs.}
\crefname{equation}{Eq.}{Eqs.}
\Crefname{equation}{Eq.}{Eqs.}
\crefname{section}{Sec.}{Secs.}
\Crefname{section}{Sec.}{Secs.}
\crefname{definition}{Def.}{Defs.}
\Crefname{definition}{Def.}{Defs.}
\crefname{algorithm}{Alg.}{Algs.}
\Crefname{algorithm}{Alg.}{Algs.}

\setlist[enumerate]{label=({\arabic*})}
\newtheoremstyle{prettydef}%
{3pt}%
{3pt}%
{}%
{}%
{\bfseries}%
{}%
{\newline}%
{\thmname{#1}\thmnumber{ #2}:\thmnote{ \textnormal{\textit{#3}}}}

\theoremstyle{prettydef}
\newtheorem{definition}{Definition}

\newcommand{\planner}{\texttt{SAST}\xspace}
\newcommand{\plannerLong}{\emph{Stable Arrangement Search Trees}}
\newcommand{\reverseproblem}{\textsc{Reverse}\xspace}
\newcommand{\transformproblem}{\textsc{Transform}\xspace}
\newcommand{\rotateproblem}{\textsc{Rotate}\xspace}
\usepackage{ifthen}
\newbool{comments}
\setbool{comments}{false}
\newcommand{\zkcom}[1]{\ifbool{comments}{{\color{red}ZK: #1}}{}}
\newcommand{\wtcom}[1]{\ifbool{comments}{{\color{blue!40}WT: #1}}{}}

\title{\LARGE \bf \myTitle{}\\ }

\author{Yiyuan Lee, Wil Thomason, Zachary Kingston, Lydia E. Kavraki%
\thanks{Department of Computer Science, Rice University, Houston, TX, 77005, USA, \texttt{\{yiyuan.lee, wbthomason, zak, kavraki\}@rice.edu}. This work was supported in part by NSF RI \#2008720 and Rice University Funds, as well as NSF Grant \#2127309 to the Computing Research Association for the CIFellows Project.}}

\begin{document}

\maketitle
\thispagestyle{empty}
\pagestyle{empty}

\begin{abstract}
	3D object reconfiguration encompasses common robot manipulation tasks in which a set of objects must be moved through a series of physically feasible state changes into a desired final configuration.
	Object reconfiguration is challenging to solve in general, as it requires efficient reasoning about environment physics that determine action validity.
	This information is typically manually encoded in an explicit transition system.
	Constructing these explicit encodings is tedious and error-prone, and is often a bottleneck for planner use.
	In this work, we explore embedding a physics simulator within a motion planner to implicitly discover and specify the valid actions from any state, removing the need for manual specification of action semantics.
	Our experiments demonstrate that the resulting simulation-based planner can effectively produce physically valid rearrangement trajectories for a range of 3D object reconfiguration problems without requiring more than an environment description and start and goal arrangements.
\end{abstract}

\section{Introduction}

Robot manipulation planning is primarily a problem of finding a sequence of \emph{valid} actions that move a set of target objects to a given goal configuration.
Actions are valid if they respect the problem's constraints, which may be task-specific (e.g., keeping a glass of water upright or maintaining dynamic stability of a stack of objects) or implicit and arising from the environment (e.g., avoiding collisions).

Most approaches to manipulation planning (e.g.,~\cite{krontiris_dealing_difficult_2015,krontiris_efficiently_solving_2016,dantam_incremental_constraint-based_2018,alami_geometrical_approach_1989,barry_manipulation_multiple_2013,toussaint_logic-geometric_programming_2015,garrett_pddlstream_integrating_2020}) rely on explicitly specified problem constraints through formal languages like Linear Temporal Logic (LTL)~\cite{he_manipulation_planning_2015} and the Planning Domain Definition Language (PDDL)~\cite{mcdermott_pddl-the_planning_1998,garrett_integrated_task_2021} or through natural language~\cite{misra_tell_me_2016}.
These manually created specifications identify both the valid, meaningful subsets of the state space and the valid transitions between these subsets.
The resulting transition system guides a search for a sequence of valid actions that perform the given task when executed by the robot.\looseness=-1

However, such specifications are onerous and error-prone to construct~\cite{thomason_counterexample-guided_repair_2021}, and may not capture the full set of possible actions.
They must not only define the valid dynamics for a problem's environment, but also be rich enough to describe a wide range of problems and goals.
Furthermore, problem specifications are not unique and the choice of specification can impact planning performance~\cite{vallati_general_approach_2018,vega-brown_admissible_abstractions_2018,dahlman_critical_assessment_2002,xia_learning_sparse_2019}.
Conversely, some manipulation planners forgo full generality for simplified problem specification and improved performance~\cite{krontiris_dealing_difficult_2015,krontiris_efficiently_solving_2016}.
These planners tend to be restricted to planar tabletop object rearrangement or similar problems~\cite{stilman_manipulation_planning_2007}.

\begin{figure}[t]
	\includegraphics[width=\linewidth]{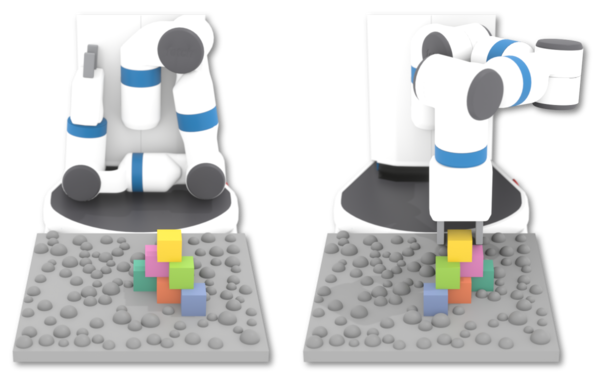}
	\caption{To rotate the pyramid, the robot must reason about the physical validity of sequences of manipulation actions.
		For example, removing cubes from the bottom of the pyramid before the cubes at the top will cause the structure to topple.
		Placing a cube at certain positions on the bumpy surface will affect the ability to transfer the other cubes.
		Explicitly specifying potential intermediate arrangements and action validity for this setting is tedious.
		Our approach leverages a physics simulator to discover the valid actions for a given arrangement and plan to reconfigure the objects.%
	}%
	\label{fig:expdemo}%
\end{figure}

We propose a middle ground: planners that can solve a broad set of classes of manipulation planning problems with no more problem specification than a typical low-level motion planning problem.
Our insight is that the necessary transition systems can be \emph{implicitly defined} through an environment simulator, reducing the manual specification burden.

This paper contributes a novel perspective on manipulation problem specification and manipulation planner design.
This perspective centers around embedding an environment simulation in a sampling-based planning algorithm as an implicit specification of a problem's valid transition system. 
In support of these ideas, we contribute
\begin{enumerate*}
	\item the \emph{arrangement space}, a novel 
	      planning space representing object arrangements and dynamically discovered low-level robot motions moving between them,
	\item \plannerLong{} (\planner), an arrangement-space planner using embedded environment simulators to discover valid action sequences and the associated low-level motions (\cref{sec:approach.sast}), and
	\item a procedure to simplify the solutions found in the arrangement space.
\end{enumerate*}
Concretely, we investigate the use of an embedded off-the-shelf physics simulator~\cite{coumans_bullet_physics_2013,lee_dart_dynamic_2018} in~\planner{} to efficiently find statically stable, collision-free states for 3D object reconfiguration problems without manually specifying action semantics.
This setting is a specific instance of the broader manipulation planning paradigm that we propose.

We demonstrate that our proposed framework can efficiently solve 3D object reconfiguration problems with physical constraints without requiring more than an environment description and start/goal configurations to specify a problem.
These results argue for the viability of a family of planners based upon implicitly simulator-defined transition systems.
\zkcom{Any discussion of how we are finding quasi-static arrangements and leveraging the capabilities of geometric planners? We talked about this briefly.}

\section{Background and Related Work}

This paper proposes a novel perspective on planning with embedded simulators that combines and extends earlier uses of simulation in robot planning and control (\cref{sec:related.work:simulation}).
\planner, an example of this perspective in practice, is an efficient sampling-based planning algorithm (\cref{sec:related.work:sbmp}) that builds upon ideas from tabletop rearrangement planning and integrated task and motion planning (\cref{sec:related.work:rearrangement}) to solve dynamically-constrained object reconfiguration problems.

\subsection{Simulation in robotics}\label{sec:related.work:simulation}

Simulation is widely used in robot control and learning.
Model-predictive control (MPC) simulates control trajectories forward through time to choose optimal inputs~\cite{garcia_model_predictive_1989}.
Recent work improves MPC performance by integrating differentiable physics simulation~\cite{heiden_interactive_differentiable_2020,heiden_neuralsim_augmenting_2021}.
Efficient simulation for training~\cite{liang_gpu-accelerated_robotic_2018,makoviychuk_isaac_gym_2021} has been core to learning-based methods for robot control~\cite{kroemer_review_robot_2021,kober_reinforcement_learning_2013,shen_acid_action-conditional_2022}, despite the challenge of translating controllers from simulation to the real world~\cite{zhao_sim-to-real_transfer_2020}.

However, as noted by~\cite{saleem_planning_selective_2020}, simulation for planning is under-studied.
Prior work combining manipulation planning and simulation restricts to specific motion primitives~\cite{zickler_efficient_physics-based_2009,zito_two-level_rrt_2012} or 2D settings~\cite{haustein_kinodynamic_randomized_2015}; we operate in a 3D workspace and dynamically discover the valid motions available to the robot via a bi-level search over object arrangements and robot motions.
~\cite{saleem_planning_selective_2020} studied efficient planning with simulators by selectively simulating actions.
~\cite{huang_parallel_monte_2022} improved the long-horizon planning efficiency of a Monte-Carlo Tree Search-based planner by integrating parallel simulation.
~\cite{agboh_robust_physics-based_2021,agboh_combining_coarse_2019} use simulators with different precisions and interleaved simulation and execution to improve manipulation planning performance and robustness.
These approaches complement~\planner.
We propose that an embedded simulator can be an effective implicit specification of a problem's constraints.

\subsection{Sampling-based motion planning}\label{sec:related.work:sbmp}
Sampling-based motion planning (SBMP) is a family of efficient robot motion planning techniques based upon constructing approximations of a robot's high-dimensional configuration space~\cite{lozano-perez_spatial_planning_1990} from sampled configurations~\cite{kavraki_probabilistic_roadmaps_1996,lavalle_planning_algorithms_2006,kuffner_rrt-connect_efficient_2000,hsu_probabilistic_foundations_2006}.
Most SBMP algorithms operate by building up a graph~\cite{kavraki_probabilistic_roadmaps_1996} or tree~\cite{lavalle_planning_algorithms_2006} of valid configurations connected by short valid motions.
\rrtconnect{}~\cite{kuffner_rrt-connect_efficient_2000} is among the fastest SBMP algorithms for many problems, due to its technique of growing two trees of valid motions, one from each of the start and the goal, toward each other using a local ``extension'' planner to control the trees' growth.
\zkcom{RRTC succeeds when the path is relatively ``straight''---i.e., the space is expansive, which is true of the arrangement space (many arrangements work, this is basically the bimonotone property we talked about). Maybe more discussion here?}
\wtcom{I'm not sure that is bimonotonicity, which (as I recall) was more about the decomposition of a problem rather than its planning space? Regardless, it might be good to note that the arrangement space is expansive.}
\planner{} adapts the high-level planning loop of \rrtconnect{} to search an expansive space of stable object arrangements (\cref{sec:approach.sast}) using a simulation-based extension planner (\cref{sec:approach.extend}).

\subsection{Object reconfiguration}\label{sec:related.work:rearrangement}

Object reconfiguration has been studied in contexts including manipulation planning~\cite{alami_geometrical_approach_1989,barry_manipulation_multiple_2013,cambon_hybrid_approach_2009,berenson_task_space_2011}, rearrangement planning~\cite{han_high-quality_tabletop_2017,krontiris_dealing_difficult_2015,krontiris_efficiently_solving_2016,shome_synchronized_multi-arm_2020}, and integrated task and motion planning (TAMP)~\cite{garrett_integrated_task_2021}.
These approaches span an axis ranging from problem specialization (i.e., planar rearrangement planners~\cite{krontiris_dealing_difficult_2015,krontiris_efficiently_solving_2016,han_high-quality_tabletop_2017}) to relative generality (i.e., full TAMP solving~\cite{dantam_incremental_constraint-based_2018,toussaint_logic-geometric_programming_2015,garrett_pddlstream_integrating_2020}).
This axis also corresponds to the relative \emph{specification effort} for each planner: a measure of the work a user must do to provide a given planner with the information it needs to operate.
Planar rearrangement planners typically only specify the desired object arrangement (as well as the environment geometry), and exploit their assumption of planar problems to find solutions faster.
TAMP solvers also rely on symbolic action specifications, mechanisms for discovering states that satisfy action preconditions, and more (e.g., explicit problem constraint specifications)~\cite{garrett_integrated_task_2021}.
We strike a balance: simulators still require manual effort to create, but are more broadly reusable across problems and domains than the specifications and samplers required by most TAMP solvers.
Simulators can also implicitly encode a more general set of constraints than most rearrangement solvers, allowing for richer problems.
Further, as progress in learning problem-specific dynamics models advances~\cite{sanchez-gonzalez_graph_networks_2018,chang_compositional_object-based_2017,battaglia_interaction_networks_2016,battaglia_simulation_engine_2013}, the effort required to create simulators for planning will decrease.

\planner, like~\cite{krontiris_efficiently_solving_2016}, relies on an arrangement-aware extension primitive to find valid action sequences.
~\cite{haustein_kinodynamic_randomized_2015} also proposes a rearrangement planner incorporating a simplified 2D physics model to evaluate a predefined set of rearrangement actions.
Similarly,~\cite{ren_rearrangement-based_manipulation_2022} explores kinodynamic planning for planar rearrangement with a focus on reacting to unexpected events during rearrangement plan execution, and using a heuristic-based task specification.
\planner{} uses full 3D physics, does not predefine motion primitives, and models dynamic constraints such as stability.
In future work, synergistically combining \planner{} with the techniques of~\cite{haustein_kinodynamic_randomized_2015,ren_rearrangement-based_manipulation_2022} could allow \planner{} to use richer non-prehensile motions for manipulating objects.

\section{Problem Formulation}

We demonstrate implicit constraint definition via embedded simulation in a specific application: 3D object reconfiguration with stability constraints, using pick-and-place actions.

Consider a 3D workspace containing movable rigid-body \emph{objects}, \(\object \in \Objects\), and a known set of posed static \emph{obstacle} geometries.
Objects have known 3D geometries and poses in \sethree.
An \emph{arrangement} assigns a pose to each object:
\begin{definition}[Arrangement]\label{def:arrangement}
	An arrangement, \(\arrangement\), prescribes a pose, \(\arrangementop{\object} \in \sethree\), to each object in the workspace.
	Denote the \emph{arrangement space}, the set of all arrangements, as \Arrangements, and let \(\arrangement \setminus o\) be arrangement \(\arrangement\) with object \(o \in \Objects\) removed from consideration.\looseness=-1
\end{definition}

Arrangements may be \emph{valid} or \emph{invalid}.
Valid arrangements are those that are both \emph{collision-free} and \emph{statically stable}.

\begin{definition}[Valid arrangement]\label{def:valid.arrangement}

	Let \collision{\cdot} be a collision test for arrangements, such that \(\collision{\arrangement} = \True\) if \arrangement{} has no objects in collision.
	Similarly, let \stable{\arrangement} be a static stability test for arrangements, such that \(\stable{\arrangement} = \True\) if \arrangement{} is statically stable after a fixed duration.

	An arrangement, \(\arrangement \in \Arrangements\), is valid if and only if \(\collision{\arrangement} = \True\) and \(\stable{\arrangement} = \True\).
	We evaluate \collision{\cdot} via a physics simulator's collision checker.
	We check \stable{\cdot} by stepping the simulator for a fixed number of time steps and verifying that all objects' displacements remain below a small heuristic threshold.\looseness=-1
\end{definition}

Let \robot{} be a robot arm with a static base and joint configuration space \CSpace~\cite{lozano-perez_spatial_planning_1990}.
The arm is capable of two classes of motion: \emph{Transit} motions move the empty end effector along a collision-free path between two workspace poses.
\emph{Transfer} motions grasp a target object and move it and the end effector to a new pose along a collision-free path~\cite{alami_geometrical_approach_1989}.
\begin{definition}[Transit motions]\label{def:transit.motion}
	A transit motion \(\transit{\arrangement, \config_i, \config_j}\) is a continuous motion of the robot arm from initial configuration \(\config_i \in \CSpace\) to \(\config_j \in \CSpace\) that is collision-free with respect to \arrangement.
\end{definition}

\begin{definition}[Transfer motions]\label{def:transfer.motion}
	A transfer motion \(\transfer{\arrangement_i, \object, \config, \config', \arrangement_j}\) is a continuous motion of the robot arm, holding object \(\object \in \Objects\), from \(\config \in \CSpace\) to \(\config' \in \CSpace\), that is collision-free with respect to \(\arrangement_i \setminus \object\).
	\(\config\) and \(\config'\) must place object \(\object\) at \(\operatorname{\arrangement_i}\lbrack\object\rbrack\) and \(\operatorname{\arrangement_j}\lbrack\object\rbrack\), respectively.
\end{definition}

\begin{figure}[t]
	\centering
	\includesvg[width=\linewidth]{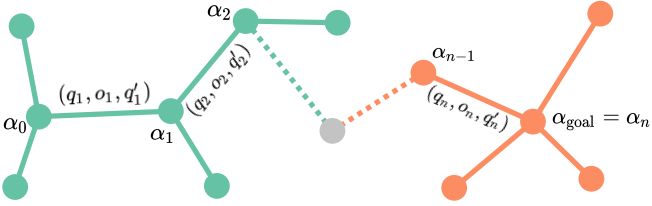}
	\caption{%
		Bidirectional search trees in the arrangement space.
		Each vertex represents a valid arrangement (\cref{def:valid.arrangement}).
		An edge \((\config_i, o_i, \config_i')\) represents a transformation between the two connected arrangements \(\arrangement_{i - 1}\) and \(\arrangement_{i}\).
		This comprises of a \transito{} motion of the robot to configuration \(\config_i\), followed by a stable \transfero{} motion of object \(o_i\) from its pose in \(\arrangement_{i - 1}\) to \(\arrangement_{i}\) by grasping \(o_i\) and moving the robot from \(\config_i\) to \(\config_i'\).
		Edges are bidirectional---one can also transform arrangement \(\arrangement_{i}\) to \(\arrangement_{i - 1}\) using a \transito{} motion to \(\config_i'\), followed by a stable \transfero{} motion of object \(o_{i}\) from its pose in \(\arrangement_{i}\) to its pose in \(\arrangement_{i - 1}\) by grasping \(o_i\) and moving the robot from \(q_i'\) to \(q_i\).
	}\label{fig:rrtc}
\end{figure}

\begin{algorithm}[t]
	\caption{\planner}\label{alg:rrtc}
	\DontPrintSemicolon
	\SetKwFunction{InitTree}{InitTree}
	\SetKwFunction{Sample}{SampleArrangement}
	\SetKwFunction{True}{True}
	\SetKwFunction{False}{False}
	\SetKwFunction{Extend}{Extend}
	\SetKwFunction{Connect}{Connect}
	\SetKwFunction{Path}{Path}
	\SetKwFunction{Swap}{Swap}
	\SetKwFunction{GetNearest}{GetNearest}

	\(\mathcal{T}_a \gets \InitTree(\arrangement_0)\)\;\label{line:rrtc_inita}
	\(\mathcal{T}_b \gets \InitTree(\arrangement_\mathrm{goal})\)\;\label{line:rrtc_initb}
	\While{\True} { \label{line:rrtc_loop}
		\(\arrangement_\mathrm{rand} \gets \Sample{}\)\; \label{line:rrtc_sample}
		\(\arrangement_\mathrm{nearest} \gets \GetNearest(\mathcal{T}_a, \arrangement_\mathrm{rand})\)\; \label{line:rrtc_nearest}
		\(\arrangement_\mathrm{new} \gets \Extend(\mathcal{T}_a, \arrangement_\mathrm{nearest}, \arrangement_\mathrm{rand}, \False) \)\; \label{line:rrtc_extend}
		\If{\(\arrangement_\mathrm{new} \not= \arrangement_\mathrm{nearest}\)} { \label{line:rrtc_ifextended}
			\If{\(\Connect(\mathcal{T}_b, \arrangement_\mathrm{new}) = \arrangement_\mathrm{new}\)}{ \label{line:rrtc_connect}
				\Return \(\Path(\mathcal{T}_a, \mathcal{T}_b)\)\; \label{line:rrtc_done}
			}
		}
		\(\Swap(T_a, T_b)\)\; \label{line:rrtc_swap}
	}
\end{algorithm}

Note that these motion classes do not predefine concrete motion primitives or actions.
We are now equipped to formally state the object reconfiguration problem:
\begin{definition}[Object Reconfiguration Problem]\label{def:reconfiguration.problem}
	Given an initial valid arrangement (\cref{def:valid.arrangement}), \(\arrangement_\mathrm{start} \in \Arrangements\), robot configuration, \(\config_\mathrm{start} \in \CSpace\), and valid goal arrangement \(\arrangement_\mathrm{goal} \in \Arrangements\), the object reconfiguration problem is to find a sequence of objects and robot configurations, \(\lbrack \config_1, \object_1, \config'_1, \ldots, \config_n, \object_n, \config'_n \rbrack\) and corresponding alternating \transito{} and \transfero{} motions such that the sequence:
	\begingroup
	\allowdisplaybreaks
	\begin{align*}
		            & \transit{\arrangement_\mathrm{start}, q_\mathrm{start}, q_1}    \\
		\rightarrow & \transfer{\arrangement_0, o_1, q_1, q_1', \arrangement_1}       \\
		\rightarrow & \cdots                                                          \\
		\rightarrow & \transit{\arrangement_{n - 1}, q_{n - 1}', q_n}                 \\
		\rightarrow & \transfer{\arrangement_{n - 1}, o_n, q_n, q_n', \arrangement_n}
	\end{align*}
	\endgroup
	is valid and
	\(\arrangement_n = \arrangement_\mathrm{goal}\), where \(\arrangement_i\) is the arrangement after executing the \(i\)-th \transfero{} motion.
\end{definition}

This problem formulation is similar to that of~\cite{krontiris_efficiently_solving_2016}, but adds a 3D workspace and consideration of stability constraints.

\section{Approach}
We propose to solve the reconfiguration problem with a bidirectional tree search algorithm, \planner{}, that operates in a given problem's arrangement space.
\planner{} resembles \rrtconnect, but operates in the arrangement space with a novel extension operator that exploits an embedded physics simulator (\cref{sec:approach.extend}) to automatically discover valid actions.

\begin{algorithm}[t]
	\caption{\texttt{Connect(}\(\mathcal{T}, \arrangement'\)\texttt{)}}
	\label{alg:connect}
	\DontPrintSemicolon

	\SetKwFunction{True}{True}
	\SetKwFunction{GetNearest}{GetNearest}
	\SetKwRepeat{Do}{do}{while}
	\SetKw{and}{and}

	\Do{\(\arrangement_\mathrm{next} \not\in \{\arrangement_\mathrm{nearest}, \arrangement'\}\)}{
		\(\arrangement_\mathrm{nearest} \gets \GetNearest(\mathcal{T}, \arrangement')\)\; \label{line:connect_nearest}
		\(\arrangement_\mathrm{next} \gets \Extend(\mathcal{T}, \arrangement_\mathrm{nearest}, \arrangement', \True)\)\;
	}
	\Return{\(\arrangement_\mathrm{next}\)}\;
\end{algorithm}

\subsection{Stable Arrangement Search Trees (\planner)}\label{sec:approach.sast}
\planner{} initializes two trees in the arrangement space, one rooted at the start arrangement, \(\arrangement_\mathrm{start}\), and the other at the goal arrangement, \(\arrangement_\mathrm{goal}\).
Vertices in these trees represent valid arrangements (\cref{def:valid.arrangement}); edges represent transformations between valid arrangements.
In this work, we consider pick-and-place transformations which move \emph{exactly} one object.
Given two valid arrangements \(\arrangement_{i - 1}\) and \(\arrangement_{i}\), a connecting edge can be described as \((\config_i, o_i, \config_i')\).
This transformation corresponds to a \transito{} motion of the robot to \(\config_i\), followed by a stable \transfero{} motion moving \(o_i\) from its pose in \(\arrangement_{i - 1}\) to \(\arrangement_{i}\) by grasping \(o_i\) and moving the robot from \(\config_{i}\) to \(\config_{i}'\).
Edges are bidirectional: the reverse transformation from \(\arrangement_{i}\) to \(\arrangement_{i - 1}\) corresponds to a \transito{} motion to \(q_i'\), followed by a stable \transfero{} motion of \(o_i\) from its pose in \(\arrangement_{i}\) to \(\arrangement_{i - 1}\) by grasping \(o_i\) and moving the robot from \(\config_{i}'\) to \(\config_{i}\).
In the arrangement space representation, a solution to a reconfiguration problem is a path of edges that connect \(\arrangement_\mathrm{start}\) to \(\arrangement_\mathrm{goal}\).\looseness=-1

Planning starts from the tree rooted at \(\arrangement_\mathrm{start}\).
Each iteration of the planning loop samples a random arrangement \(\arrangement_\mathrm{rand}\) and finds its closest neighbor, \(\arrangement_\mathrm{nearest}\) in the current tree (\cref{alg:rrtc}, \cref{line:rrtc_sample,line:rrtc_nearest}).
This is done via spatial lookup on a GNAT~\cite{brin_neighbor_search_1995} with arrangement distance defined as the summed \sethree{} distance\footnote{\sethree{} distance is the sum of the Euclidean distance of the translational components and the angular distance of the rotational components.} between the respective poses of each object in the two arrangements.
\planner{} then attempts to \texttt{Extend} the tree from \(\arrangement_\mathrm{nearest}\) toward \(\arrangement_\mathrm{rand}\) by growing a sequence of edges according to~\cref{alg:extend}.
If the resulting sequence of edges is non-empty (\cref{alg:rrtc}, \cref{line:rrtc_ifextended}), we try to \texttt{Connect} the other tree to the terminal vertex of the extended trajectory (\cref{alg:rrtc}, \cref{line:rrtc_connect}).
This is done (\cref{alg:connect}) by repeatedly extending the closest arrangement on the other tree to the terminal vertex, until either the connection succeeds or until the extension fails.
If connection succeeds, \planner{} has found a solution and terminates.
Otherwise, it swaps the trees and repeats the planning loop.\looseness=-1

\begin{figure}[t]
	\includegraphics[width=\linewidth]{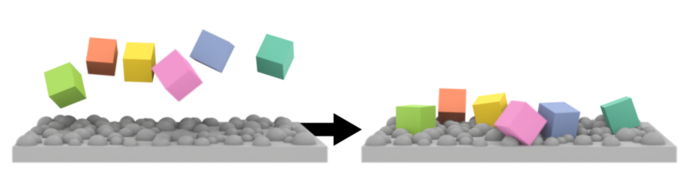}
	\caption{Stable arrangement sampling.
	The pose of each object is uniformly sampled within workspace bounds; rejection sampling ensures no object intersection.
	The dynamics of the world are stepped until the arrangement stabilizes, that is, zero displacement over a sufficient number of steps.
	}\label{fig:sa}
\end{figure}

\subsection{Sampling stable arrangements}

The \texttt{SampleArrangement} subroutine (\cref{fig:sa}) samples a valid arrangement for use with \texttt{Extend}.
Here, we leverage the embedded physics simulator to find stable arrangements.
First, \texttt{SampleArrangement} picks uniform-random 3D poses for each object within the workspace bounds, using rejection sampling to ensure that the objects do not intersect.
Then, it simulates the dynamics of the arrangement forward for a fixed number of small timesteps, checking at fixed intervals if the objects have maintain zero displacement since the previous interval.
If so, the arrangement resulting from the applied dynamics is kinematically valid and statically stable, and is returned as a result.
Otherwise, this process repeats until a valid sample is found.
\texttt{SampleArrangement} is easy to parallelize---our implementation of \planner{} uses multiple threads to sample stable arrangements.

Uniform-random initial pose sampling trades off performance for ease of specification, avoiding the specialized samplers used by TAMP solvers to find states on low and zero-measure state manifolds.

\subsection{Generating valid transformation actions}\label{sec:approach.extend}

The \Extend subroutine (\cref{alg:extend}) searches for a sequence of valid edges that transform a given arrangement \(\arrangement\) of \(k\) objects into a target arrangement \(\arrangement'\).
A major contribution of our work is to use an embedded physics simulator in \Extend to reason about the validity of these transformations.
The simulator allows us to treat the physics of the environment as an implicit specification of the valid transformation actions from any state.
We also ensure that a valid transformation has a valid instantiation with the robot by motion planning for its associated \transito{} and \transfero{} motions.

\begin{algorithm}[t]
	\caption{\texttt{Extend(}\(\mathcal{T}, \arrangement, \arrangement', connectTarget\)\texttt{)}}
	\label{alg:extend}
	\DontPrintSemicolon
	\SetKwFunction{RandomObjectOrder}{RandomObjectOrder}
	\SetKwFor{ForParallel}{for}{do parallel}{end for}
	\SetKwFunction{InitTree}{InitTree}
	\SetKwFunction{Sample}{SampleArrangement}
	\SetKwFunction{True}{True}
	\SetKwFunction{Extend}{Extend}
	\SetKwFunction{Connect}{Connect}
	\SetKwFunction{Path}{Path}
	\SetKwFunction{Swap}{Swap}
	\SetKwFunction{Copy}{Copy}
	\SetKwFunction{IKOp}{SampleGraspConfs}
	\SetKwFunction{ParentEdge}{PrecedingConf}
	\SetKw{if}{if}
	\SetKw{then}{then}
	\SetKw{continue}{continue}
	\SetKw{or}{or}
	\SetKw{donot}{not}
	\SetKw{and}{and}
	\SetKw{invalid}{invalid}
	\SetKw{valid}{valid}
	\SetKwFunction{AddVertex}{AddVertex}
	\SetKwFunction{AddEdge}{AddEdge}
	\SetKwFunction{Transit}{TRANSIT}
	\SetKwFunction{Transfer}{TRANSFER}

	\(o_1, \dots, o_k \gets \RandomObjectOrder{}\)\; \label{line:extend_random}
	\(\arrangement_\mathrm{cur} \gets \Copy(\arrangement)\)\; \label{line:extend_copy}
	\For{\(i\gets 1\) \KwTo k}{ \label{line:extend_loop}

		\(\arrangement_\textrm{next} \gets (\arrangement_\mathrm{cur}[o_i] \gets \arrangement'[o_i])\)\; \label{line:extend_next}

		\tcp{\textcolor{gb}{Sample grasp configurations.}}
		\(q_i, q_i' \gets \IKOp(\arrangement_\mathrm{cur}[o_i], \arrangement'[o_i])\)\;\label{line:extend_ik}

		\tcp{\textcolor{gb}{Perform checks.}}
		\if \(\collision{\arrangement_\mathrm{next}} = \False\) \continue\; \label{line:extend_check3}
		\if \(\stable{\arrangement_\mathrm{cur} \setminus o_i} = \False\) \continue\;\label{line:extend_check4}
		\if \(\stable{\arrangement_\mathrm{next} \setminus o_i} = \False\) \continue \;\label{line:extend_check5}
		\(q_\mathrm{prev}' \gets \ParentEdge(\arrangement_\mathrm{cur})\)\; \label{line:extend_parentedge}
		\if \invalid \Transit{\(\arrangement_\mathrm{cur}, q_\mathrm{prev}', q_i\)} \or
		\invalid \Transfer{\(\arrangement_\mathrm{cur}, o_i, q_i, q_i', \arrangement_\mathrm{next}\)} \continue\; \label{line:extend_check7}

		\tcp{\textcolor{gb}{Connect target, or create edge.}}
		\eIf{\(\arrangement_\mathrm{next} = \arrangement'\) \and \(connectTarget\)}{ \label{line:extend_checkconnect1}
			\(q_\mathrm{next}' \gets \ParentEdge(\arrangement')\)\; \label{line:extend_childedge}
			\If{\valid \Transit{\(\arrangement_\mathrm{next}, q_i', q_\mathrm{next}'\)}}{ \label{line:extend_checkconnect2}
				\(\mathcal{T}.\AddEdge(\arrangement_\mathrm{cur}, \arrangement', (q_i, o_i, q_i'))\)\;\label{line:extend_connectadd}
				\Return{\(\arrangement'\)}\;\label{line:extend_connectreturn}
			}
		}{
			\(\mathcal{T}.\AddVertex(\arrangement_\textrm{next})\)\;\label{line:extend_add1}
			\(\mathcal{T}.\AddEdge(\arrangement_\mathrm{cur}, \arrangement_\textrm{next}, (q_i, o_i, q_i'))\)\;\label{line:extend_add2}
			\(\arrangement_\mathrm{cur} \gets \arrangement_\textrm{next}\)\;\label{line:extend_add3}
		}
	}
	\Return{\(\arrangement_\mathrm{cur}\)}\;\label{line:extend_return}
\end{algorithm}

\Extend starts by selecting a random order to move the objects\footnote{We choose a random order for simplicity, but could substitute a more sophisticated permutation selector for performance.} and setting the current arrangement, \(\arrangement_\mathrm{cur}\) to the given start arrangement, \(\arrangement\).
It then tries to move each object in the chosen order to its target position in the given target arrangement, \(\arrangement'\), while maintaining stability of the other objects.\looseness=-1

\begin{figure*}[t]
	{\centering
		\begin{tabularx}{\linewidth}{YYYYYY}
			{\includegraphics[width=1.15\linewidth]{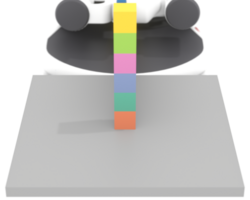}}   &
			{\includegraphics[width=1.15\linewidth]{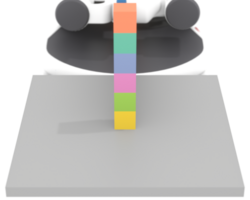}}    &
			{\includegraphics[width=1.15\linewidth]{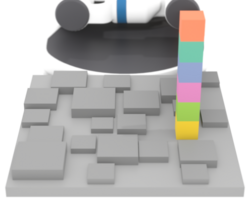}} &
			{\includegraphics[width=1.15\linewidth]{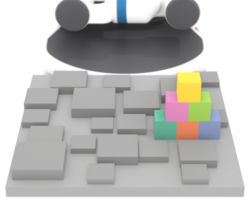}}  &
			{\includegraphics[width=1.15\linewidth]{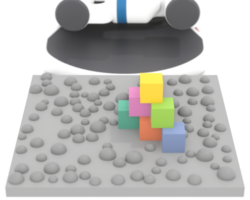}}    &
			{\includegraphics[width=1.15\linewidth]{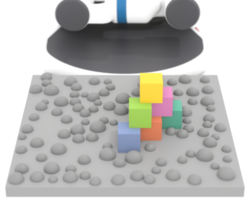}}
			\\
			\mbox{\footnotesize{(a) \reverseproblem Start}}                                            &
			\mbox{\footnotesize{(b) \reverseproblem Goal}}                                             &
			\mbox{\footnotesize{(c) \transformproblem Start}}                                          &
			\mbox{\footnotesize{(d) \transformproblem Goal}}                                           &
			\mbox{\footnotesize{(e) \rotateproblem Start}}                                             &
			\mbox{\footnotesize{(f) \rotateproblem Goal}}
		\end{tabularx}
	}
	\caption{Test problems used in our experiments.
		(a-b) \reverseproblem: The robot starts with a stack of \(6\) cubes and must re-stack them on the same base location in reversed order.
		(c-d) \transformproblem: The robot must transform a stack of \(6\) cubes into a pyramid, centered at the same location. The table is covered in tiles of random height and size, which constrain the feasible intermediate arrangements.
		(e-f) \rotateproblem: The robot is given a \textit{diagonal} pyramid of cubes and must manipulate the cubes into another pyramid with cubes stacked cross-diagonally. The table is covered in bumps of random size and location, which prevent the cubes from being placed flat in intermediate arrangements and make it more difficult to find stable arrangements.
		In all problems, the robot starts with its arm tucked in (\cref{fig:expdemo}). Across runs, the \(x\) and \(y\) positions of the structures to reconfigure, as well as the tiles and bumps, are randomized.}\label{fig:expproblems}
\end{figure*}

For each object, \(o_i\), \Extend creates a new arrangement \(\arrangement_\mathrm{next}\) equal to \(\arrangement_\mathrm{cur}\) with \(o_i\) at its pose in \(\arrangement'\) and samples collision free robot configurations grasping \(o_i\)'s pose in \(\arrangement_\mathrm{cur}\) and at \(\arrangement_\mathrm{next}\), using the same grasp.
Then, it checks that:
\begin{enumerate*}
	\item \(\arrangement_\mathrm{next}\) is collision-free,
	\item \(\arrangement_\mathrm{cur} \setminus o_i\) is stable, allowing \(o_i\) to be moved, and
	\item \(\arrangement_\mathrm{next} \setminus o_i\) is also stable, allowing \(o_i\) to be moved in the \emph{reverse} transformation
\end{enumerate*}.
If these conditions are met, \Extend attempts to find a valid \transito{} motion between the preceding configuration of \(\arrangement_\mathrm{cur}\)\footnote{If \(\arrangement_\mathrm{cur} = \arrangement_\mathrm{start}\), we select \(q_0\) as \(q_\mathrm{prev}'\).
	If \(\arrangement_\mathrm{cur} = \arrangement_\mathrm{goal}\), we skip this check since there is no constraint on the robot configuration at the goal.} and the sampled grasp for \(o_i\)'s pose in \(\arrangement_\mathrm{cur}\), and a valid \transfero{} motion between the sampled grasps for \(o_i\)'s pose in \(\arrangement_\mathrm{cur}\) and \(\arrangement_\mathrm{next}\), respectively, using a standard motion planner (\cref{alg:extend}, \cref{line:extend_parentedge,line:extend_check7}).
These motions are considered infeasible if the sub-planner fails to find a solution within a predefined timeout.

If \Extend finds the requisite \transito{} and \transfero{} motions, then it either adds the discovered edge to the \emph{current} tree (\cref{alg:extend}, \cref{line:extend_add1,line:extend_add2,line:extend_add3}) and continues with the next object and \(\arrangement_\mathrm{cur} = \arrangement_\mathrm{next}\), or attempts to connect the newly-reached arrangement to the \emph{other} tree (\cref{alg:extend}, \cref{line:extend_checkconnect1,line:extend_childedge,line:extend_checkconnect2,line:extend_connectadd}) if \(\arrangement_\mathrm{next}\) is the target arrangement and a connection is desired.
In the latter case, \Extend returns the target arrangement (\cref{alg:extend}, \cref{line:extend_connectreturn}); otherwise, it iterates through the remaining objects and returns the last reached arrangement.\looseness=-1

\subsection{Solution simplification}\label{sec:approach.simplification}

Although unnecessary for completeness, \planner{} applies heuristic simplifications to solutions to improve their quality.

If an object \(o\) has been moved twice along a solution trajectory, one of these motions may be unnecessary.
We can remove the first motion by altering the second motion to move \(o\) starting from the first motion's starting pose.
Similarly, we can remove the second motion by altering the first motion to move \(o\) to the second motion's ending pose.
Both cases modify the pose of \(o\) in the arrangements between the first and second motions.
This requires recomputing the grasps and planning motions for these intermediate arrangements to validate the altered arrangement trajectory.
In a third case, motions may also be removed if the pickup and placement locations of the object are exactly the same.

\planner{} iterates through these three simplification cases on solutions, rechecking for stability and recomputing the \transito{} and \transfero{} motions after each modification to ensure that the solution remains feasible.
This simplification process continues until no potentially redundant actions remain.
Note that this heuristic set is non-exhaustive and does not guarantee optimal motions.\looseness=-1

\begin{figure*}[t]
	{\centering
		\begin{tabularx}{\linewidth}{YYYY}
			{\includegraphics[width=\linewidth]{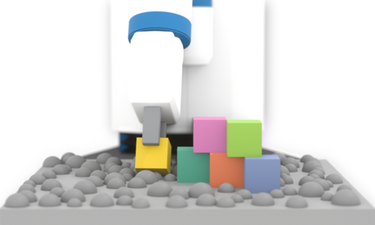}} &
			{\includegraphics[width=\linewidth]{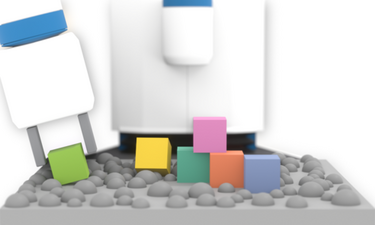}} &
			{\includegraphics[width=\linewidth]{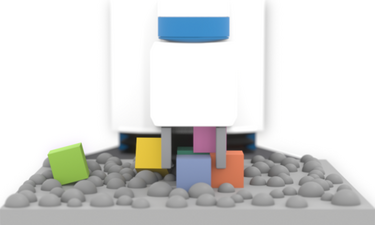}} &
			{\includegraphics[width=\linewidth]{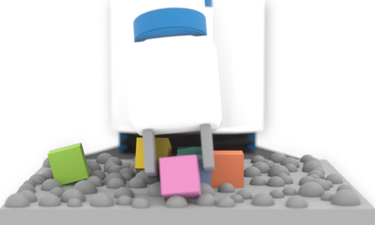}}   \\
			{\includegraphics[width=\linewidth]{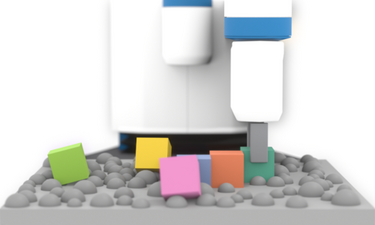}} &
			{\includegraphics[width=\linewidth]{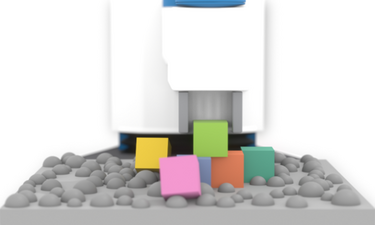}} &
			{\includegraphics[width=\linewidth]{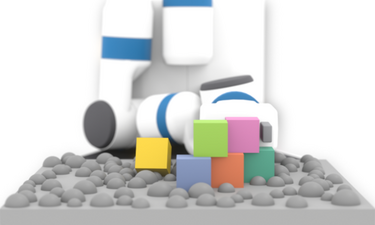}} &
			{\includegraphics[width=\linewidth]{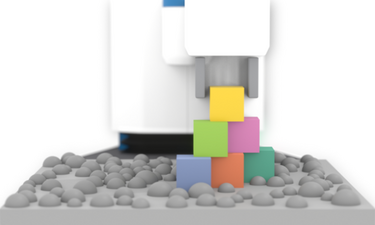}}   \\
		\end{tabularx}
	}
	\caption{Sequence of actions in a computed solution after simplification for the \rotateproblem problem.
		The robot is able to identify actions that remove the blocks on top before removing those at the bottom.
		This ensures that blocks are not removed from the bottom that will cause those stacked on top to topple.
		The sequence shown achieves the minimum possible length for the given problem.}\label{fig:expsimplified}
\end{figure*}

\section{Experiments}

We evaluate \planner{} on a set of 3D tabletop rearrangement problems (\cref{fig:expproblems})---\reverseproblem (a-b), \transformproblem (c-d), and \rotateproblem (e-f).
These problems involve using a single-arm manipulator to reconfigure cubes from one 3D structure to another. They require reasoning about the physical constraints between the cubes, as well as with the environment.
The solutions are non-trivial, in that the robot must choose and move the objects through intermediate arrangements to achieve its goal.
In addition, some problems contain obstacles such as tiles and bumps which complicate the validity of actions.
Grounding these details in order to apply contemporary approaches would be tedious and challenging.
\zkcom{Why are these problems interesting? Point out interesting features: requiring intermediate locations, hard to specify in PDDL, etc.}
\wtcom{Also, are we adding the experiments with multi-level tables?}

\subsection{Implementation}

We use DART~\cite{lee_dart_dynamic_2018} as our embedded physics simulator and plan \transito{} and \transfero{} motions via Robowflex~\cite{kingston_robowflex_robot_2022} with the Open Motion Planning Library (OMPL)~\cite{sucan2012open}.
All experiments ran on an AMD 5900x desktop CPU with \(12\) cores at \(4.8\) GHz, using \(6\) parallel threads for sampling stable arrangements and inverse kinematics.

\subsection{Planning performance}

\begin{table}[t]
	\centering
	\begin{tabularx}{\linewidth}{Yccc}
		\toprule
		                & \reverseproblem & \transformproblem & \rotateproblem \\
		\midrule
		Success Rate    & 0.96 (0.03)     & 0.96 (0.03)       & 1.00 (0)       \\
		Solve Time (s)  & 16.5 (0.9)      & 55.0 (5.6)        & 61.5 (6.6)     \\
		Solution Length & 25.6 (0.7)      & 23.7 (0.8)        & 15.2 (0.5)     \\
		Num. Nodes      & 44.1 (2.4)      & 86.2 (9.6)        & 57.4 (6.0)     \\
		\bottomrule
	\end{tabularx}
	\caption{Run-time metrics of \planner{} across the test problems.
		\textit{Success rate} refers to the proportion of runs where \planner{} finds a solution within the given time limit.
		For successful runs,
		\textit{Solve Time} refers to the time taken to find a solution;
		\textit{Solution Length} refers to the solution length, in terms of the number of objects moved;
		\textit{Num. Nodes} refer to the total number of tree vertices created by \planner{} during search.
		Mean values are shown, with standard error shown in parentheses.}\label{expres:planning}
	\vspace{-0.2cm}
\end{table}

We applied \planner{} to each test problem for \(50\) trials with a maximum timeout of \(300\) seconds per trial. In each trial, we randomize the start and goal positions of the structure to rearrange, together with the the obstacle positions.

\Cref{expres:planning} shows that \planner{} was almost always able to find a solution within the stipulated time limit.
Solution times were also reasonable, taking not more than a minute per successful run despite having to invoke the simulator repeatedly for collision and static stability checking, and having to integrate the low-level motion planning of the \transito{} and \transfero{} motions.
The sizes of the search trees, in terms of the number of nodes, were also small, indicating that a sparse coverage of arrangements was sufficient to identify a solution.

Across each problem and trial, we only had to specify the geometry and positions of the obstacles (steps and bumps) and the start and goal arrangement poses of the objects. 
This highlights the strength of our approach in using the physics simulator to automatically derive action validity without requiring any manual, explicit specification.

Solution lengths, however, often require about twice the optimal number of steps.
This is because \planner, like \rrtconnect, is non-optimizing.\looseness=-1

\subsection{Simplification performance}

The results in~\cref{expres:planning} do not use the solution simplification heuristics of~\cref{sec:approach.simplification}.
\Cref{expres:simplification} shows the results of applying these heuristics to the solutions found by each successful run, indicating that solution simplification usually terminated within \(40\) seconds.
Most of the additional time comes from rechecking for stability and replanning for the low-level \transito{} and \transfero{} motions, required whenever two actions moving the same object merge.
This is done up to \(\mathcal{O}(n^3)\) times in the number of actions in the initial solution.

Simplification usually decreased solution length by roughly half, reaching or coming close to the optimal solution length.
\Cref{fig:expsimplified} shows an example of a simplified \rotateproblem{} solution.\looseness=-1

\begin{table}[t]
	\centering
	\begin{tabularx}{\linewidth}{Yccc}
		\toprule
		                           & \reverseproblem & \transformproblem & \rotateproblem \\
		\midrule
		Simplify Time (s)          & 26.5 (1.4)      & 41.8 (2.7)        & 28.0 (2.0)     \\
		Simplified Solution Length & 12.1 (0.1)      & 12.0 (0.1)        & 8.0 (0.1)      \\
		Improvement (\%)           & 51.5 (1.2)      & 46.8 (1.7)        & 44.9 (1.7)     \\
		\bottomrule
	\end{tabularx}
	\caption{Solution simplification results.
		\textit{Simplify Time} is the time taken for the simplification procedure to terminate.
		\textit{Simplified Solution Length} is the length of the simplified solutions, in terms of the number of objects moved.
		\textit{Improvement} is the percentage of the original solution length that simplification eliminated.
		Mean values are shown, with standard error in parentheses.\looseness=-1%
	}\label{expres:simplification}
	\vspace{-0.2cm}
\end{table}

\subsection{How important is integrating motion planning?}

\planner{} verifies the feasibility of each \transito{} and \transfero{} motion by planning a valid trajectory in the robot's configuration space for the motion.
To investigate the impact of this integrated verification, we conducted an ablation experiment by removing these low-level feasibility checks.
We find solutions in terms of sequences of object arrangements and object grasps, assuming that the transformations between arrangements are feasible.
After finding a full solution, we attempt to compute a motion plan for each of the associated low-level motions to check solution validity.

\Cref{expres:primitiveablation} shows a substantial drop in solution feasibility when low-level motion checks are skipped.
Indeed, \transito{} motions require checking that an object is reachable with a given grasp without the manipulator colliding with the other objects; \transfero{} motions require checking that an object can be pulled away without the object or the manipulator intersecting with the other nearby objects. The problems we consider have environment obstacles (table, tiles, and bumps) that do not interfere much with the robot's motion---in more constrained environments, such as in a cupboard or drawer, we could expect feasibility to worsen.

\begin{table}[t]
	\centering
	\begin{tabularx}{\linewidth}{Yccc}
		\toprule
		                  & \reverseproblem & \transformproblem & \rotateproblem \\
		\midrule
		w/ Motion Checks  & 1.00 (0)        & 1.00 (0)          & 1.00 (0)       \\
		w/o Motion Checks & 0.60 (0.07)     & 0.42 (0.07)       & 0.56 (0.07)    \\
		\bottomrule
	\end{tabularx}
	\caption{Solution feasibility with and without low-level motion checking during the arrangement tree search.
		We show the mean proportion of solutions with feasible low-level motions; the standard error is in parentheses.\looseness=-1%
	}\label{expres:primitiveablation}
	\vspace{-0.2cm}
\end{table}

\section{Discussion}

This work contributes a novel perspective on manipulation planning that embeds a simulator to implicitly encode the valid actions and states for a problem domain.
We demonstrate this perspective for 3D object reconfiguration planning, where we are able to efficiently find statically stable object configurations that would otherwise be onerous to specify.

\planner{} currently uses random sampling and extension to grow the arrangement space graph, but informed approaches like Expansive Space Trees~\cite{hsu_path_planning_1997} or Monte Carlo Tree Search~\cite{huang_parallel_monte_2022} may discover solutions faster.
SBMP advances such as biased samplers~\cite{hsu_bridge_test_2003,Lee2022} and optimizing planners~\cite{karaman_sampling-based_algorithms_2011,gammell_asymptotically_optimal_2021,strub_aitstar_eitstar_2022,janson2015fast} may also complement embedded-simulator planning.

Embedded-simulator planning is broadly applicable outside object reconfiguration, which poses several directions for future work.
How can we use simulators to encode constraints beyond stability, such as orientation or contact dynamics?
Similarly, what are the precise requirements for an embedded simulator?
For some problems, precise physics simulation may be unnecessary; for others, non-standard physics can encode problem constraints.
Further, how well do plans found via embedded simulation transfer to the real world?\looseness=-1

Finally, we wish to explore richer uses of the embedded simulator, including combining differentiable simulation with optimization techniques, to broaden the manipulation problem classes that we can efficiently solve.

\newpage
\balance
\printbibliography{}

\end{document}